\documentclass[11pt]{article}

\usepackage[final]{acl}
\usepackage{amsmath}
\usepackage{times}
\usepackage{latexsym}

\usepackage[T1]{fontenc}

\usepackage[utf8]{inputenc}

\usepackage{microtype}

\usepackage{inconsolata}

\usepackage{graphicx}

\usepackage{booktabs}

\usepackage{xcolor}

\title{Subgoal-aware Memory Graph for GUI Agent}

\author{
  Dezhao Luo \and Daniel Beechey \and Derek Yuen \and Weilin Luo\thanks{Corresponding author. Email: \texttt{luoweilin3@huawei.com}} \\
  Huawei Noah's Ark Lab
}

\begin{document}
\maketitle

\begin{abstract}
GUI agents aim to automate complex user tasks by interacting with mobile interfaces, but effective decision-making in long-horizon settings requires reusing prior experience rather than relying solely on the environment observations. Existing memory-based methods typically store experience either at the single-action level or as entire trajectories. Single-step actions are often too fine-grained to carry meaningful semantics, since the success of an individual action does not necessarily imply progress towards the overall goal. In contrast, whole trajectories are often too coarse-grained and difficult to reuse across different tasks.

We propose a subgoal-aware memory graph that distils past trajectories into structured, reusable evidence at the level of subgoals. We define a subgoal as consecutive actions sharing a single intent, the granularity at which progress toward the task becomes meaningful. To reduce redundancy and support cross-task reuse, we organise subgoal experiences as nodes in a graph, where related memories can be merged, extended, and retrieved based on the environment observation. On SPA-Bench, our method improves success rate by 11.11\% over the baseline.
\end{abstract}

\begin{figure*}[t]
    \centering
    \includegraphics[width=\textwidth]{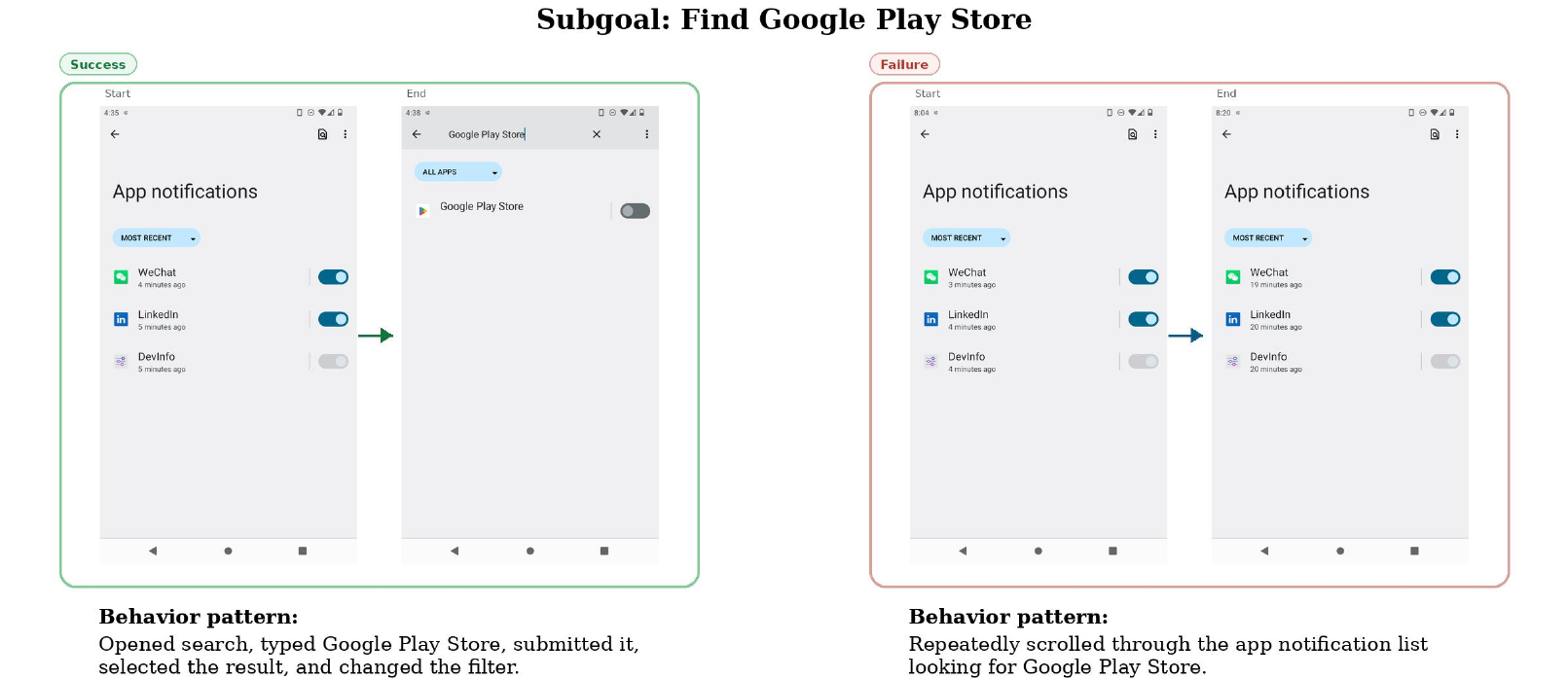}
    \caption{The same subgoal can succeed (left) or fail (right) depending on the underlying behaviour. Subgoal granularity exposes progress toward the user goal that is invisible at the single-action level.}
    \label{fig:method_overview}
\end{figure*}

\section{Introduction}
GUI agents aim to complete complex user goals by interacting with graphical user interfaces, from web pages~\citep{zheng2024seeact,deng2023mind2web,zhou2024webarena} to mobile phone screens, such as ordering items on a shopping app, setting up an alarm, or navigating settings~\citep{zhang2024appagent,wang2024mobileagent,hong2024cogagent,rawles2023aitw,rawles2024androidworld}. In benchmarks like SPA-Bench~\citep{chen2024spabench}, a single task can require more than 30 GUI actions across several screens. In such long-horizon settings, relying solely on the environment observation is insufficient because many subgoals recur across tasks and apps, and without reusing prior interaction experience, an agent must re-explore the same workflows from scratch~\citep{xie2025guiexplorer,zhu2026hymem}.

A straightforward solution is to equip the agent with an episodic memory it can query at inference time~\citep{packer2023memgpt,park2023generative}. Existing memory-based methods, however, typically store experience either at the single-action level~\citep{xie2025guiexplorer} or as entire trajectories~\citep{zhu2026hymem}, and both granularities are problematic on mobile GUIs. Single-step actions are too fine-grained to carry semantics on task progress, since whether a tap or swipe ``succeeds'' locally tells the agent little about whether the overall goal is being approached. Whole trajectories, on the other hand, are too coarse-grained, since a full episode, as stored in representations such as HyMEM~\citep{zhu2026hymem}, is tied to a specific task path, making it rarely reusable across tasks that share intermediate steps but diverge in specifics.

We argue for an intermediate abstraction level, namely \emph{subgoals}, which we define as contiguous sequences of actions that jointly express a coherent sub-intent. Unlike raw actions, subgoals reflect progress toward the goal. As illustrated in Figure~\ref{fig:method_overview}, the same subgoal can either succeed or fail depending on the underlying behaviour, a distinction that is invisible at the single-action level. Unlike whole trajectories, subgoals transfer naturally across related tasks, since many tasks share the same underlying subgoals.

Building on this granularity, we organise past experience into a subgoal-level memory graph. Each raw trajectory is first segmented into a chain of subgoals. Rather than storing the full action sequence, we represent each subgoal as a transition between its boundary visual states (the starting and ending screenshots), annotated with a subgoal description, a success label, a behaviour summary of the underlying actions, and an effect description obtained by comparing the boundary states. Boundary states are then embedded and merged across trajectories when visually similar, so that each subgoal becomes a directed edge between two shared memory nodes. The resulting graph consolidates reusable visual states and the semantic transitions between them, and can be incrementally extended with new trajectories.

On the SPA-Bench~\citep{chen2024spabench} English single-app level-3 split, our subgoal-aware memory graph achieves an improvement of 11.11\% over a memory-free baseline with the same Qwen3-VL-8B-Instruct~\citep{qwen2025qwen3vl} backbone. 


\begin{figure*}[t]
    \centering
    \includegraphics[width=\textwidth]{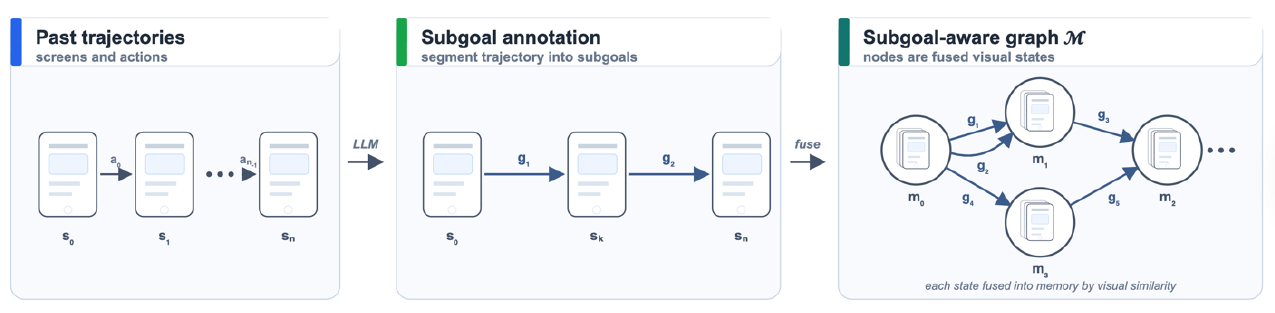}
    \caption{Overview of the proposed subgoal-aware memory framework. Past trajectories are segmented into annotated subgoals and fused into a graph over shared visual states.}
    \label{fig:method}
\end{figure*}

\section{Related Work}
\paragraph{GUI agents.}
App agents interpret natural-language goals and operate smartphones through actions such as tapping, scrolling, and text entry~\citep{rawles2023aitw,rawles2024androidworld,wang2025distrl,luo2025vimo,beechey2026darwin}. One line of work fine-tunes vision-language models to obtain end-to-end GUI agents~\citep{hong2024cogagent,qin2025uitars,tang2025asl,christianos2025lightweight}, achieving strong in-distribution accuracy at the cost of substantial training. A complementary line of work keeps the underlying LLM frozen. AppAgent~\citep{zhang2024appagent} maintains a hand-curated, per-app knowledge base, whereas Mobile-Agent~\citep{wang2024mobileagent} introduces visual perception tools to predict actions. Our work targets this frozen-LLM setting, where the agent's capabilities are shaped by what is supplied through its context rather than by weight updates.

\paragraph{Memory for GUI agents.}
Memory mechanisms aim to give GUI agents access to relevant past experience at inference time, so that decisions are informed by what worked or failed in earlier rollouts. Existing memory designs for GUI agents cluster at two representational granularities. GUI-Explorer~\citep{xie2025guiexplorer} mines per-step \emph{(observation, action, outcome)} triples through autonomous exploration, and such single-action units carry little information on task progress, and a memory of them quickly accumulates redundant entries without exposing reusable structure. HyMEM~\citep{zhu2026hymem} organises continuous trajectory embeddings into a graph and EchoTrail-GUI~\citep{li2025echotrail} retrieves whole successful episodes as in-context demonstrations. However, full trajectories are tied to the path of a single task. Our work targets the intermediate \emph{subgoal} granularity that these designs leave underexplored. Subgoals are contiguous action segments that mark progress toward task completion and recur across different tasks. We organise them in a graph whose nodes are shared visual states, rather than per-task or per-action entries.

\section{Method}
\label{sec:method}

Our goal is to equip a GUI agent with a memory that distils past interaction experience into reusable, subgoal-level guidance. Let a trajectory $\tau = (s_0, a_0, s_1, a_1, \dots, a_{n-1}, s_n)$ denote a sequence alternating between GUI screenshots $s_i$ and executed actions $a_i$. Given a corpus $\mathcal{T} = \{\tau_1, \dots, \tau_m\}$ of such trajectories, we construct a memory graph $\mathcal{M}$ that the agent can query during task execution: given the environment screenshot $s_q$, the memory returns a small set of subgoal-level guidance items that are appended to the agent's prompt. We instantiate $\mathcal{M}$ as a directed graph whose nodes correspond to clusters of visually similar GUI states and whose edges encode subgoals observed in past trajectories, each edge connecting the state at which a subgoal began to the state at which it ended. The construction and retrieval pipeline comprise three stages: subgoal extraction (Section~\ref{sec:method:subgoals}), graph construction (Section~\ref{sec:method:graph}), and memory retrieval (Section~\ref{sec:method:retrieval}).

\subsection{Subgoal Extraction and Annotation}
\label{sec:method:subgoals}

\paragraph{Segmentation.}
We segment each trajectory into a chain of \emph{subgoals}, where a subgoal spans one or more consecutive actions that jointly realise a coherent sub-intent, for example, \emph{``open the search bar and submit a query''} or \emph{``add the selected pair of sunglasses to the basket''}. Segmentation is performed by an LLM that takes the full action sequence as input and returns segment boundaries. We keep the first and last screenshots of each segment as its visual anchors and discard the intermediate frames, as the boundary screenshots already capture the start and end states relevant for downstream retrieval.

\paragraph{Subgoal metadata.}
Each extracted subgoal $g \in \mathcal{G}$ is represented as a typed tuple
\[
\begin{aligned}
g = \langle\, & s^{\mathrm{from}},\, s^{\mathrm{to}};\\
              & \texttt{text},\, \texttt{effect},\, \texttt{behaviour},\, \texttt{success}\, \rangle
\end{aligned}
\]
whose fields are populated as follows.
\begin{itemize}
  \item $s^{\mathrm{from}}, s^{\mathrm{to}}$: the boundary screenshots, used as visual anchors during graph construction.
  \item \texttt{text}: a natural-language summary of the subgoal's intent.
  \item \texttt{effect}: a one-sentence description of the observed visual change from $s^{\mathrm{from}}$ to $s^{\mathrm{to}}$.
  \item \texttt{behaviour}: a concise summary of the underlying action sequence, highlighting the dominant interaction pattern.
  \item \texttt{success} $\in \{\texttt{yes}, \texttt{no}, \texttt{uncertain}\}$: a subgoal-level success label assigned by an LLM judge.
\end{itemize}

\subsection{Subgoal-aware Memory Graph Construction}
\label{sec:method:graph}

Given the annotated subgoals from Section~\ref{sec:method:subgoals}, we now organise them into a persistent, queryable memory structure. The goal is to support behaviour reuse across tasks: when the agent encounters a state it has seen before, it should be able to retrieve which subgoals were previously attempted from that state, whether they succeeded, and through what action behaviour. We cast this as a directed graph whose nodes are GUI states and whose edges are subgoal transitions. Construction proceeds as follows: (1) merging visually similar screenshots into shared nodes so that experience accumulates rather than fragments; (2) attaching each subgoal as a directed edge carrying its full annotation; and (3) deduplicating redundant edges within each node to keep the graph compact.

\paragraph{Environment state merging.}
We map the boundary screenshots of each annotated subgoal, $s^{\mathrm{from}}$ and $s^{\mathrm{to}}$, into a shared memory-node space according to visual similarity. For each such screenshot $s$, we compute its image embedding and compare it against the existing memory nodes. If the nearest node $m_j$ satisfies $\cos(s, m_j) \geq \tau$, where $\cos(\cdot, \cdot)$ denotes cosine similarity between image embeddings, the screenshot is absorbed into $m_j$. Otherwise, a new memory node is instantiated. Each node maintains its constituent screenshots together with a running mean embedding used for subsequent matching.

\paragraph{Directed subgoal edges.}
After environment state merging, each annotated subgoal becomes a directed edge from its source memory node to its target memory node. The edge stores the subgoal intent, the observed visual effect between the endpoint screenshots, the success label, and the behaviour pattern of the executed action sequence. Crucially, the graph preserves both successful and failed attempts from each state, so retrieval can later provide failed alongside successful examples.

\paragraph{Build-time deduplication.}
To keep the memory compact as more trajectories are added, we remove repeated outgoing subgoals within each source node. Two outgoing subgoals are treated as duplicates only when they point to the same target node, carry the same success label, and describe the same intent, i.e., the cosine similarity between text embeddings of their intent descriptions exceeds $\theta$. When a duplicate is detected, the later edge is discarded. Subgoals that differ in target, outcome, or intent are kept as separate evidence.

\subsection{Memory Retrieval}
\label{sec:method:retrieval}


At each decision step, retrieval is triggered with the current screenshot $s_q$. We embed $s_q$ and compare it against the embedding maintained at each memory node, and keep the top-$k$ nearest nodes by cosine similarity. For each retrieved node, we collect all of its outgoing subgoal edges, and both successful and failed subgoals are returned, each tagged with its outcome, so that the agent can avoid known failure modes as well as imitate known successes. The retrieved subgoals are then formatted into a short memory block and added to the agent's prompt.

\section{Experiments}
\label{sec:exp}

\subsection{Setup}
\label{sec:exp:setup}

\paragraph{Benchmark and Environment.}
We run experiments on cloud-based Android phone instances controlled through ADB. We use the English single-app level-3 split of SPA-Bench~\citep{chen2024spabench}, the more difficult tier of the benchmark and a small number of apps that cannot run on cloud phone instances are excluded from evaluation. This leaves 13 tasks in our evaluation. We use Qwen3-VL-8B-Instruct \cite{qwen2025qwen3vl} as the base model. Both image and text embeddings used in memory construction and retrieval are computed by Qwen3-VL-Embedding-8B~\citep{qwen2025qwen3vlembedding}. We set both the visual similarity threshold $\tau$ and the text similarity threshold $\theta$ to $0.9$, and retrieve the top $k = 1$ nearest memory node at each decision step. In our experiment, memory construction requires 4 GPU-hours, and each retrieval incurs an additional latency of approximately 0.5 seconds.

\subsection{Results}
\label{sec:exp:main}
Table~\ref{tab:main} reports the per-attempt success rate (SR), averaged over three independent runs. With the same backbone and setup, our subgoal-graph memory improves success by 11.11\% over the memory-free baseline, suggesting that retrieving past experience is useful for long-horizon mobile tasks. Moreover, neither action-level nor task-level memory yields meaningful gains over the baseline. 

We further ablate the graph structure by replacing it with a flat memory store, where subgoals from different environments are not organised through shared visual-state nodes. This reduces SR to 30.77\%. We argue that the graph structure allows redundant memories to be merged across trajectories, keeping the retrieved context precise.

\begin{table}[t]
\centering
\small
\setlength{\tabcolsep}{16pt}  
\renewcommand{\arraystretch}{1.15}  
\begin{tabular}{llc}
\toprule
Memory   & Structure & SR (\%) \\
\midrule
\textit{None}     & --    & 34.19 \\
\midrule
Subgoal  & Flat  & 30.77 \\
Action   & Graph & 38.46    \\
Task     & Graph & 36.75    \\
\midrule
Subgoal (Ours)  & Graph & $\mathbf{45.30}$ \\
\bottomrule
\end{tabular}
\caption{Per-attempt success rate. Our subgoal-graph memory outperforms both the no-memory baseline and flat / alternative-granularity variants.}
\label{tab:main}
\end{table}
\section{Conclusion}

We proposed a subgoal-aware memory graph for mobile GUI agents. The core idea is to store past experience at the level of subgoals, rather than raw actions or whole trajectories. To make these subgoals reusable across trajectories, we fuse visually similar environment states into shared memory nodes and attach subgoal transitions between them. This allows the agent to see what has been tried from similar states and what outcome each subgoal produced. At inference time, the agent retrieves subgoals from the graph and uses them as guidance for the current task. This provides a simple and practical memory structure for mobile interaction.

\section*{Limitations}
Our method assumes that GUI environments can be approximated by a discrete state space, such that screenshots can be meaningfully clustered into a compact set of shared memory nodes. This assumption is reasonable for typical mobile applications with relatively stable layouts, but becomes limited in highly dynamic interfaces, such as video players, where consecutive states change continuously, and visual similarity is a noisier proxy for semantic equivalence.

\bibliography{references}

\end{document}